\documentclass{article}
\usepackage[english]{babel}
\usepackage[utf8]{inputenc}
\usepackage{johd}

\hypersetup{
    colorlinks,
    linktoc=none,
    linkcolor=blue,
    citecolor=blue,
    urlcolor=blue
}

\newcommand*{\footnotemarkcolor}{black}
\makeatletter
\renewcommand*{\@makefnmark}{\hbox{\@textsuperscript{%
   \color{\footnotemarkcolor}\normalfont\@thefnmark}}}
\makeatother

\title{Improving Classification Performance With Human Feedback: \\ Label a few, we label the rest}

\author{Eden Chung, Liang Zhang, Katherine Jijo, Thomas Clifford, Natan Vidra}

\date{} %leave blank

\begin{document}

\maketitle

\begin{abstract} 
\noindent 
In the realm of artificial intelligence, where a vast majority of data is unstructured, obtaining substantial amounts of labeled data to train supervised machine learning models poses a significant challenge. To address this, we delve into few-shot and active learning, where are goal is to improve AI models with human feedback on a few labeled examples. This paper focuses on understanding how a continuous feedback loop can refine models, thereby enhancing their accuracy, recall, and precision through incremental human input. By employing Large Language Models (LLMs) such as GPT-3.5, BERT, and SetFit, we aim to analyze the efficacy of using a limited number of labeled examples to substantially improve model accuracy. We benchmark this approach on the Financial Phrasebank, Banking, Craigslist, Trec, Amazon Reviews datasets to prove that with just a few labeled examples, we are able to surpass the accuracy of zero shot large language models to provide enhanced text classification performance. We demonstrate that rather than needing to manually label millions of rows of data, we just need to label a few and the model can effectively predict the rest.
  
\end{abstract}

%---------------------------------------------%

\section{Introduction}
In the world of AI, a significant challenge is handling massive amounts of data. Only 15 percent of this data is structured, while the rest, a surprising 85 percent, is unstructured. For AI/ML models to work effectively, they usually need large sets of data that are labeled, but getting such data is challenging.
\\
\\
Traditional AI methods rely heavily on millions of rows to train models. This means pairing inputs (like pictures or text) with the correct labels. The big question is: How can we gather this massive amount of labeled data efficiently? Right now, businesses have to choose between leaving their data unlabeled (which is risky) or manually (AI Assisted) labeling it. While AI tools like Labelbox, Heartex, Datasaur and Prodigy can help with labeling, they aren't perfect. Training data often necessitates the expertise of subject-matter professionals, such as doctors, legal analysts, and financial analysts, for labeling.  This manual data labeling is tedious, time-consuming, and costly, but particularly, when business requirements change, requiring manual labeling over and over again is not sustainable. Furthermore, manual labeling does not even ensure the correctness of the data; in fact, it can oftentimes be incorrect, which is one of the limitations of relying solely on manual labeling. Even state-of-the-art data sets, such as MNIST and ImageNet can have incorrectly labeled data \footnote{Curtis Northcutt and Anish Athalye and Jonas Mueller. Label Errors in ML Test Sets. \url{https://labelerrors.com}, Accessed on 2023-12-23.}. On the other hand, AI models such as GPT3 and Claude, although surely very helpful, can still hallucinate and return false data. Therefore, an approach that integrates human expertise with AI can accelerate labeling processes while ensuring accuracy, as well as minimizing errors, especially for diverse data types from different industries. 
\\
\\
One such approach that combines human power with AI is programmatic labeling \footnote{Programmatic labeling. \url{https://snorkel.ai/programmatic-labeling/}, Accessed on 2023-12-23.
}. Instead of manually labeling each data point one by one, in programmatic labeling, the user inputs labeling functions, capturing the reasoning behind the labeling, which can then be generalized to larger amounts of unlabeled data through AI. This has been a great step forward, allowing for scalability and adaptability. However, there are several limitations to programmatic labeling. Since creating labels is generalizing data, the model may struggle with ambiguous cases or nuanced cases, and labeling functions may not be able to capture these subtle patterns that a human might be able to. In addition, as mentioned before, humans can make errors in labeling, and if these errors occur in the labeling functions, these errors will propagate throughout the dataset, instead of being limited to just one incorrect data point. Finally, as manual labeling is, programmatic labeling is still relatively resource-intensive. 
\\
\\
Novel NLP research in Large Language Models and few shot learning has changed the way that data labeling is done. GPT-3, introduced in ``Language Models Are Few Shot Learners" \footnote{Tom B. Brown and Benjamin Mann and Nick Ryder and Melanie Subbiah, et al. Language Models are Few-Shot Learners, 2020. arXiv:2005.14165.}, 2020, demonstrated the abilities for LLMs to learn with minimal data. LaMDA \footnote{Romal Thoppilan, Daniel De Freitas, Jamie Hall, Noam Shazeer, Apoorv Kulshreshtha, et al. LaMDA: Language Models for Dialog Applications, 2022. arXiv:2201.08239.} and PaLM \footnote{Aakanksha Chowdhery, Sharan Narang, Jacob Devlin, Maarten Bosma, Gaurav Mishra et al. PaLM: Scaling Language Modeling with Pathways, 2022. arXiv:2204.02311.
}, significantly contributed to the enhancement of language models. In 2021, ``Want To Reduce Labeling Cost? GPT-3 Can Help" was published \footnote{Shuohang Wang and Yang Liu and Yichong Xu and Chenguang Zhu and Michael Zeng. Want To Reduce Labeling Cost? GPT-3 Can Help, 2021. arXiv:2108.13487}. The paper highlighted that employing labels produced by GPT-3 is notably more economical, both in terms of computational resources and potentially time, compared to acquiring labels from human experts. Furthermore, even with these cost benefits, models trained using GPT-3.5-generated labels demonstrated comparable performance to those trained on human-provided labels.
\\
\\
From this research, combining AI and human power called ``few-shot learning" became the practical. Instead of labeling millions of data points, we can work with just a few thousand. This approach lets AI learn from a small amount of labeled data and then make educated guesses for the rest.  Large language models such as GLaM \footnote{Nan Du, Yanping Huang, Andrew M. Dai, Simon Tong, et al. GLaM: Efficient Scaling of Language Models with Mixture-of-Experts, 2022. arXiv:2112.06905.}, Flamingo \footnote{Jean-Baptiste Alayrac, Jeff Donahue, Pauline Luc, Antoine Miech, et al. Flamingo: a Visual Language Model for Few-Shot Learning, 2022. arXiv:2204.14198.}, and the ``Alexa Teacher Model" \footnote{Jack FitzGerald, Shankar Ananthakrishnan, Konstantine Arkoudas, Davide Bernardi, et al. Alexa Teacher Model: Pretraining and Distilling Multi-Billion-Parameter Encoders for Natural Language Understanding Systems. In Proceedings of the 28th ACM SIGKDD Conference on Knowledge Discovery and Data Mining (KDD ’22), August 2022. \url{http://dx.doi.org/10.1145/3534678.3539173}, DOI: 10.1145/3534678.3539173.}, all demonstrate the expanding capabilities of AI models in learning efficiently from limited data. These models have set new benchmarks in the field, showcasing the power of few-shot learning in diverse applications, from language understanding to visual recognition and beyond.
\\
\\
In 2022, Hugging Face announced SetFit, an efficient framework for few-shot fine-tuning of Sentence Transformers, which allows high accuracy with little labeled data \footnote{Lewis Tunstall and Nils Reimers and Unso Eun Seo Jo and Luke Bates and Daniel Korat and Moshe Wasserblat and Oren Pereg. Efficient Few-Shot Learning Without Prompts, 2022. arXiv:2209.11055.}. Few-shot learning is a technique that allows a model to utilize just a minimal number of examples to guide the model. By incorporating feedback from industry experts and integrating few-shot learning techniques, we can significantly improve the overall accuracy of the model with a small amount of labeled data provided by the industry expert. By combining state-of-the-art transformers with few-shot learning techniques, the system learns in real time, assisting data annotators with the labeling process.

%---------------------------------------------%

\section{Background}

LLM's have evolved vastly over time, with this mainly being due to  the architecture behind the models. Several main architectural methods have been used; we will discuss Recurrent Neural Networks (RNN) and Transformers.

\subsection{Recurrent Neural Networks (RNN)}
Recurrent Neural Networks (RNN), the earlier architecture behind LLMs, was a concept thought up of in 1986, but the architecture for the model was finally built only in 1997. RNN has evolved over time to 3 distinct phases: Overloaded Single Memory (Vanilla RNN), Multiple Gate Memories (LSTM), and Encoder-Decoder architecture \footnote{Chen Yanhui. A Battle Against Amnesia: A Brief History and Introduction of Recurrent Neural Networks. \url{https://towardsdatascience.com/a-battle-against-amnesia-a-brief-history-and-introduction-of-recurrent-neural-networks-50496aae6740}, Accessed on 2023-12-17.}.

\subsection{Transformers}
With technology constantly changing and deep learning models improving over time, RNN began to be replaced with a new architectural model – transformers. First introduced in 2016, they have since revolutionized the field of LLMs and NLP. Beginning with BERT in 2018, models have continued to evolve with GPT in 2020 and LaMDA in 2021 \footnote{Ashish Vaswani and Noam Shazeer and Niki Parmar and Jakob Uszkoreit and Llion Jones and Aidan N. Gomez and Lukasz Kaiser and Illia Polosukhin. Attention Is All You Need, 2023. arXiv:1706.03762.}, proving that LLMs keep getting better and better, and it is clear that these models will only get more impressive with time.
\\
\\
Transformers are now the core architecture behind many NLP models such as Chat GPT and Bard. Instead of having to consider words sequentially, as RNN models do, they can instead analyze words simultaneously. As a result, transformers are very good at adapting, so many LLMs exist today with many different use cases.

% \autocite{Dang}.

\begin{figure}[h!]
\centering
\caption{Transformer architecture}
\includegraphics[scale=0.23]{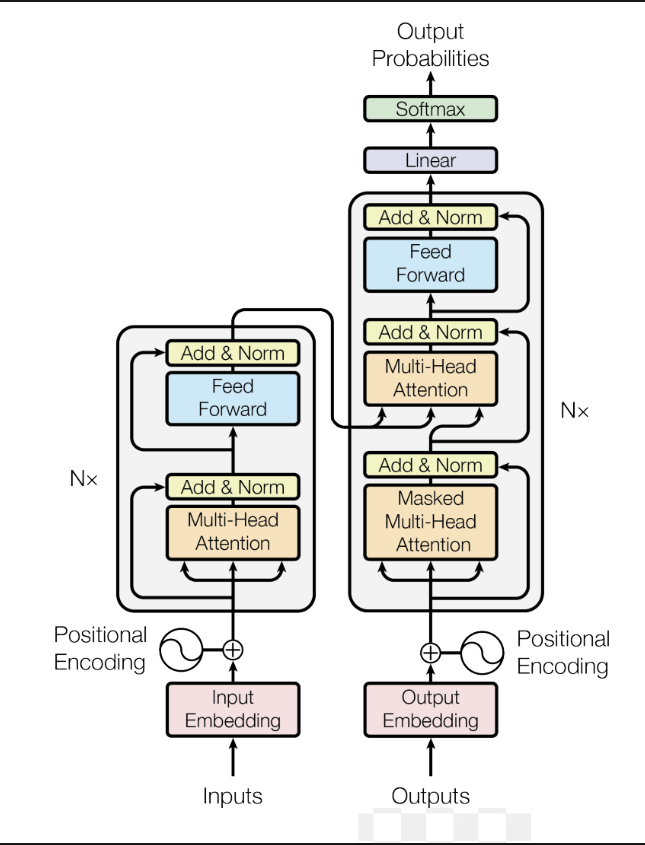}
\end{figure} 
\footnote{Liz McQuillan. Deep Learning 101: What Is a Transformer and Why Should I Care? \url{https://www.saltdatalabs.com/blog/deep-learning-101/what-is-a-transformer-and-why-should-i-care}, Accessed on 2023-12-17}

%---------------------------------------------%

\section{Process and Work}
In order to attempt to improve the accuracy of ML models, a human feedback approach is introduced, enhancing text classification accuracy. Similar to providing students with clear examples for better understanding, the strategy involves fine-tuning models through human feedback or providing more labeled examples.
\\
\\
In order to evaluate whether or not few-shot learning is really effective or not, in this study, a continual test and evaluation loop is used. The process consists of continually testing a datasets accuracy, precision, and recall but varying the number of labeled data points inputted. Starting with 10 data labeled data points given, after each iteration, human feedback is applied. 10 incorrect answers are chosen and then fed into the model, giving it the correct answer instead, and this cycle repeats. So, on each iteration, the number of labeled data points increases by 10.

\subsection{Data Sets}

In our machine learning research, we utilized six distinct datasets for various analyses. The Amazon reviews dataset consisted of a training set with 6,001 rows and a testing set of 2,001 rows, encompassing labels such as Excellent', Very Good', Neutral', Good', and Bad'. The banking dataset, on the other hand, had a training data of 200 rows and a testing set of 2,000 rows, with 77 distinct labels like cash received', fiat currency support', and pin blocked'. The Craigslist dataset was structured with 201 rows in its training set and 1,001 rows in its testing set, covering categories like phone, furniture, housing, electronics, and car, with labels categorized into ABBR, ENTY, DESC, HUM, LOC, and NUM. The financial phrasebank dataset had 4,850 rows of data with categories positive, negative and neutral. Lastly, the Trec dataset offered both a coarse label set of 6 labels and a fine label set of 50 detailed labels such as expression abbreviated, animals, organ of body, color, invention, book, and other creative pieces. This Trec is consistent with 5,452 rows in training and 500 rows in testing.

\begin{figure}[h!]
\centering
\caption{Sample of the Trec Data Set}
\includegraphics[scale=0.4]{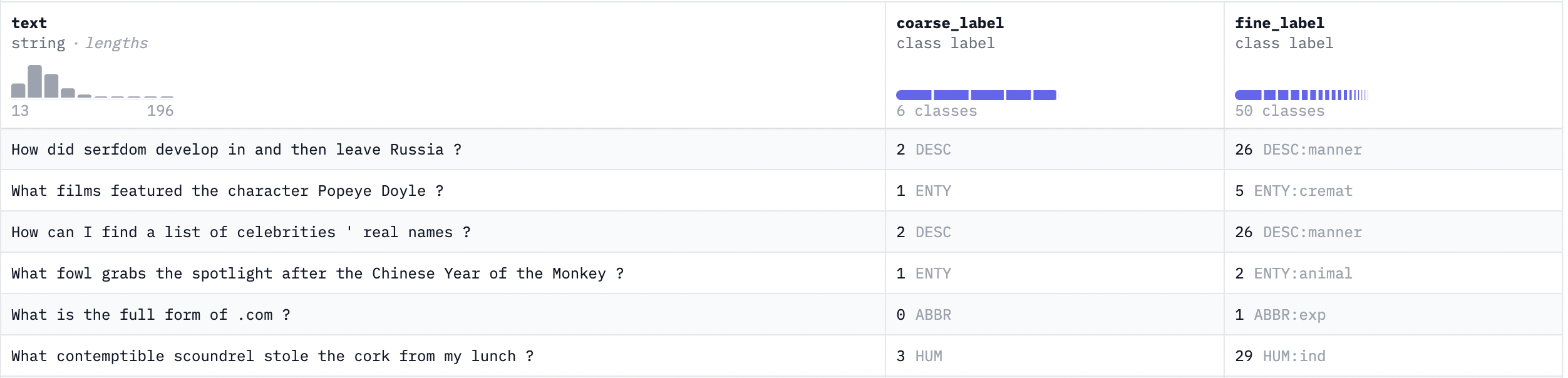}
\end{figure} 

\begin{figure}[h!]
\centering
\caption{Sample of the Finance Phrasebank Set}
\includegraphics[scale=0.4]{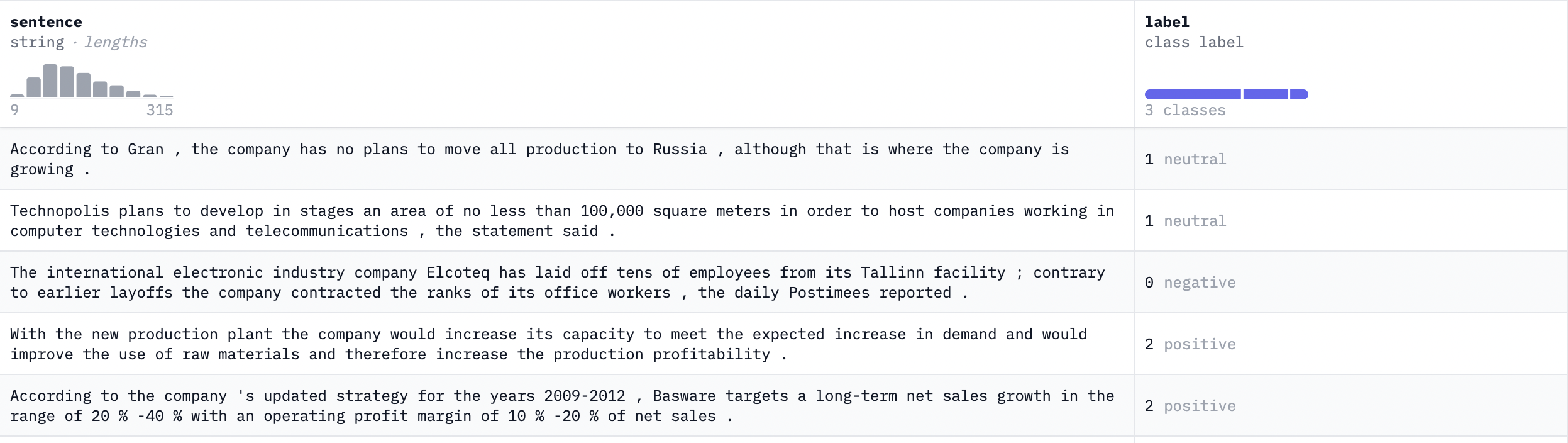}
\end{figure}

%---------------------------------------------%

\section{Approach}  

In text classification, key metrics play a pivotal role. The probability metric indicates the confidence level of a model's predictions, while the entropy metric measures the significance or unpredictability of a prediction within the dataset. Higher entropy suggests that the predicted label has an impact on the overall classification or decision-making process. To initiate our process, we utilize zero-shot models like Claude, GPT, BERT, and SETFIT to derive initial predictions from the provided data. For clarity on our predictions for a spam versus non-spam dataset, before doing anything we have to examine the zero-shot model predictions.
\newline

\begin{table}[!htbp]
\centering
\caption{Zero shot model predictions}
\begin{tabular}{|p{6cm}|l|l|l|}
\hline
\textbf{Text Body} & \textbf{Predicted} & \textbf{Probability} & \textbf{Entropy} \\ \hline
Important notice: Your package has been delivered. & not spam & 0.65 & 0.88 \\ \hline
Dear customer, Your account balance is low. & not spam & 0.58 & 0.82 \\ \hline
Hi, How are you doing? Let's catch up soon. & not spam & 0.58 & 0.75 \\ \hline
Urgent notice: Last chance to update your personal information. & spam & 0.92 & 0.51 \\ \hline
Hi there, You have won a free vacation! Claim now! & spam & 0.95 & 0.42 \\ \hline
Congratulations! You've won a million dollars! & spam & 0.97 & 0.36 \\ \hline
\end{tabular}
\label{table:spam_detection}
\end{table}

\noindent The first step in active learning from human feedback is to generate a prioritized list of edge cases. These edge cases highlight instances where the model's predictions demonstrate uncertainty or reduced confidence. Once this list is established, human evaluators can review and assign the appropriate labels. In Table 2, we can see that the actual label is not spam. By iterating through this process, the model is once again organized based on entropy and probability values. Upon this second evaluation, you'll notice a decrease in entropy and an increase in probability scores, highlighting the model's enhanced confidence and refined predictions.

\begin{table}[!htbp]
\centering
\caption{Spam Classification Results}
\begin{tabular}{|p{6.5cm}|l|l|l|l|}
\hline
\textbf{Text} & \textbf{Actual Label} & \textbf{Predicted} & \textbf{Probability} & \textbf{Entropy} \\ \hline
Important notice: Your package has been delivered. & not spam & not spam & 1.0 & 0 \\ \hline
Urgent notice: Last chance to update your personal information. &  & not spam & 0.88 & 0.28 \\ \hline
Dear customer, Your account balance is low. &  & spam & 0.70 & 0.76 \\ \hline
Hi, How are you doing? Let's catch up soon. & & not spam & 0.65 & 0.68 \\ \hline
Congratulations! You've won a million dollars! & & spam & 0.92 & 0.22 \\ \hline
Hi there, You have won a free vacation! Claim now! & & spam & 0.80 & 0.25 \\ \hline
\end{tabular}
\label{table:spam_classification}
\end{table}

\noindent We iterate on this process again. The model once again is sorted by entropy, we provide the row / edge case with the highest potential for impact.

\begin{table}[!htbp]
\centering
\caption{Iterated Model Predictions - First Iteration}
\begin{tabular}{|p{6.5cm}|l|l|l|l|}
\hline
\textbf{Text} & \textbf{Actual Label} & \textbf{Predicted} & \textbf{Probability} & \textbf{Entropy} \\ \hline
Important notice: Your package has been delivered. & not spam & not spam & 1.0 & 0 \\ \hline
Dear customer, Your account balance is low. &  & spam & 0.70 & 0.76 \\ \hline
Hi, How are you doing? Let's catch up soon. & & not spam & 0.65 & 0.68 \\ \hline
Urgent notice: Last chance to update your personal information. &  & not spam & 0.88 & 0.28 \\ \hline
Hi there, You have won a free vacation! Claim now! & & spam & 0.80 & 0.25 \\ \hline
Congratulations! You've won a million dollars! & & spam & 0.92 & 0.22 \\ \hline
\end{tabular}
\label{table:iterated_model_first}
\end{table}

\noindent Here, we see the sentence \textit{Dear customer, Your account balance is low.} which has the highest entropy of the non-labeled data points. We label this row as spam, largely due to the fact that the sentence is referring to account balance. Notice how this row of text data, with the corresponding label, is added to the annotation history once we confirm the annotation. After this second label, the models predictions, uncertainty and entropy is adjusted once more.

\begin{table}[!htbp]
\centering
\caption{Iterated Model Predictions - Second Iteration}
\begin{tabular}{|p{6.5cm}|l|l|l|l|}
\hline
\textbf{Text} & \textbf{Actual Label} & \textbf{Predicted} & \textbf{Probability} & \textbf{Entropy} \\ \hline
Important notice: Your package has been delivered. & not spam & not spam & 1.0 & 0 \\ \hline
Dear customer, Your account balance is low. & spam & spam & 1.0 & 0 \\ \hline
Hi, How are you doing? Let's catch up soon. & & not spam & 0.80 & 0.62 \\ \hline
Urgent notice: Last chance to update your personal information. & & spam & 0.86 & 0.18 \\ \hline
Hi there, You have won a free vacation! Claim now! & & spam & 0.75 & 0.20 \\ \hline
Congratulations! You've won a million dollars! & & spam & 0.88 & 0.18 \\ \hline
\end{tabular}
\label{table:iterated_model_second}
\end{table}

\noindent Notice how over time, the entropy is decreasing as the model is becoming more stable as more human labels are added. After a few labels, the entropy is able to stabilize, and no more labels are needed (the labels impact the accuracy less than is worth the effort).

%---------------------------------------------%

\section{Evaluating Method}  
We primarily employed the BERT, SEFIT, and GPT-3.5 Turbo models for our evaluation. Initially, we tasked these models with predicting outcomes on the testing set. This approach helped us identify areas where the models exhibited weakness or lower confidence. For any incorrect predictions made by the models, we adjusted the labels for 10 rows to reflect the correct classifications. We then fine-tuned the models based on this corrected data, simulating human feedback, and measured accuracy, precision, and recall on the testing set after feedback. Subsequently, we incorporated these 10 rows into the training set and excluded them from the testing set. We repeated this process iteratively while measuring their accuracy, precision, and recall. We continued this iterative approach up to 150 labels.

\section{Results}
After fine-tuning the models with human feedback and additional labeled examples, we observed a consistent improvement in accuracy across different datasets. Notably, our experiments revealed that incorporating targeted training in areas where the model is weak played a pivotal role. This iterative approach allowed the model to gradually enhance its proficiency in handling specific domains. To see the raw results and code in more detail, please access it \href{https://github.com/Whiteii/Anote-Text-Classification}{here}.

\begin{figure}[h!]
\centering
\caption{Amazon Dataset}
\includegraphics[scale=0.5]{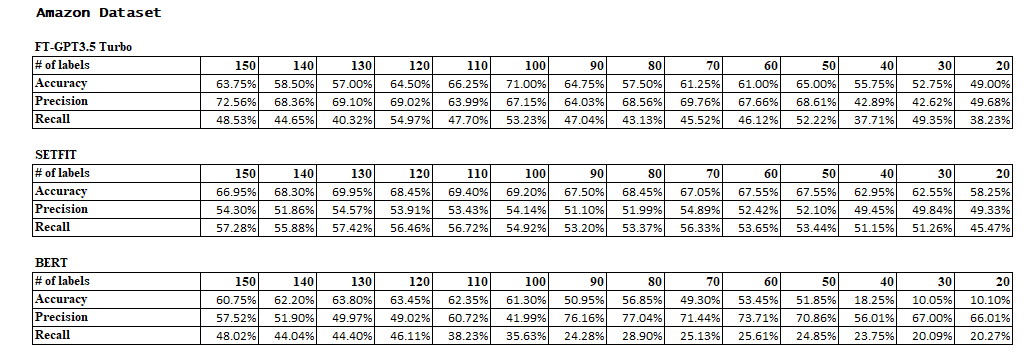}
\end{figure} 

\begin{figure}[h!]
\centering
\caption{Amazon Dataset Plot}
\includegraphics[scale=0.4]{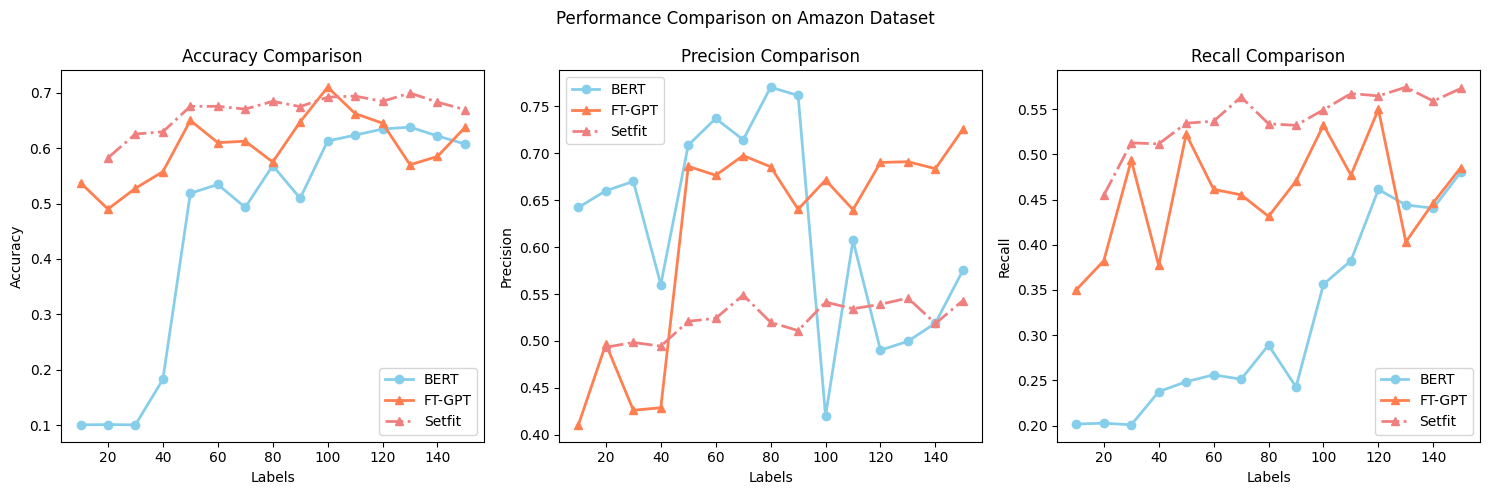}
\end{figure} 

\begin{figure}[h!]
\centering
\caption{Banking Dataset}
\includegraphics[scale=0.5]{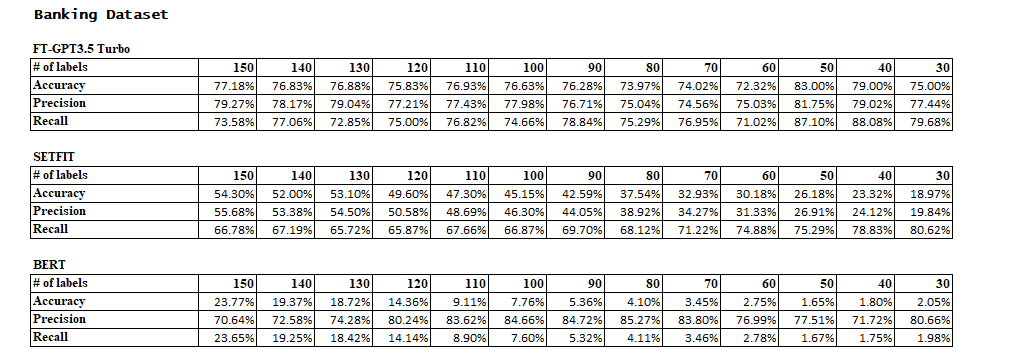}
\end{figure} 

\begin{figure}[h!]
\centering
\caption{Banking Dataset Plot}
\includegraphics[scale=0.4]{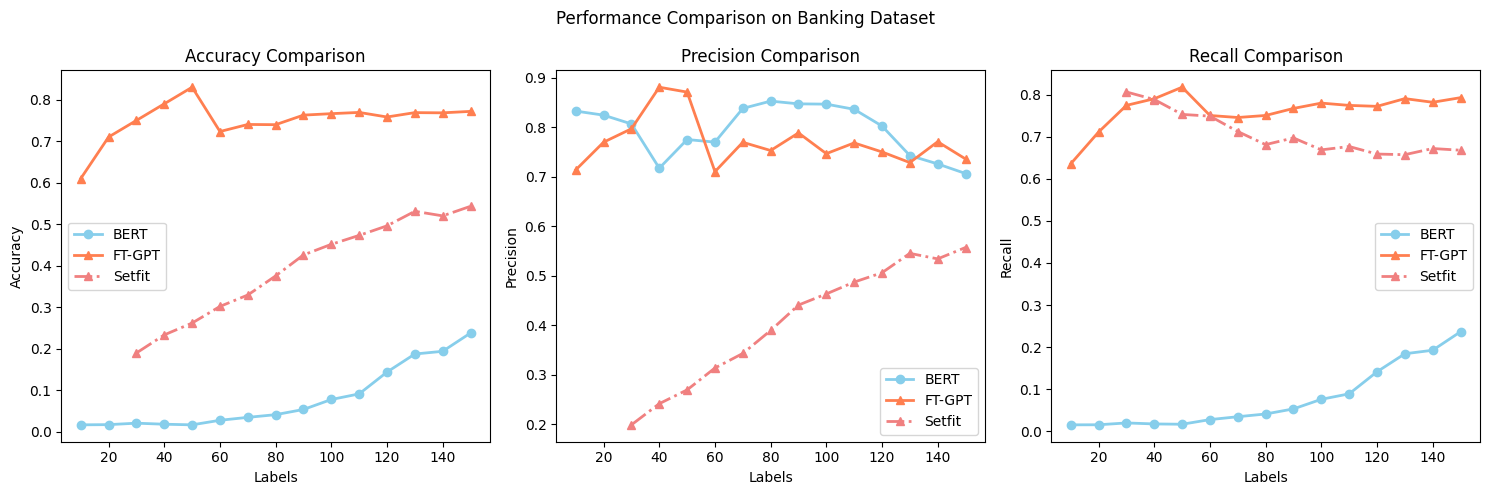}
\end{figure} 

\begin{figure}[h!]
\centering
\caption{Craigslist Dataset}
\includegraphics[scale=0.5]{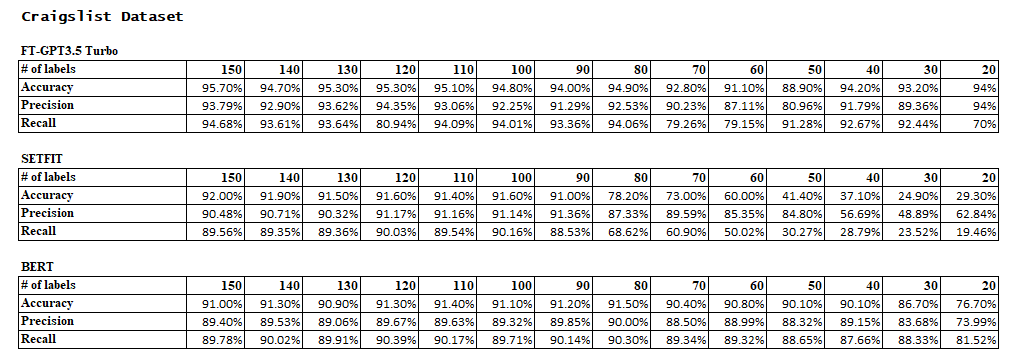}
\end{figure} 

\begin{figure}[h!]
\centering
\caption{Craigslist Dataset Plot}
\includegraphics[scale=0.4]{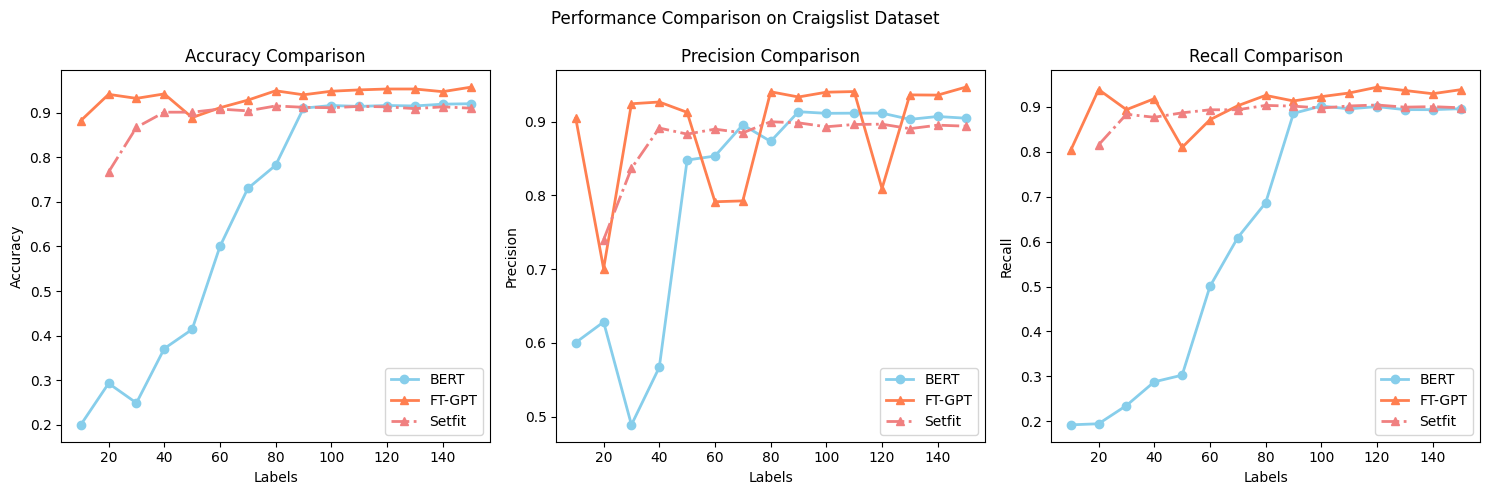}
\end{figure}

\begin{figure}[h!]
\centering
\caption{Financial Phrasebank Dataset}
\includegraphics[scale=0.5]{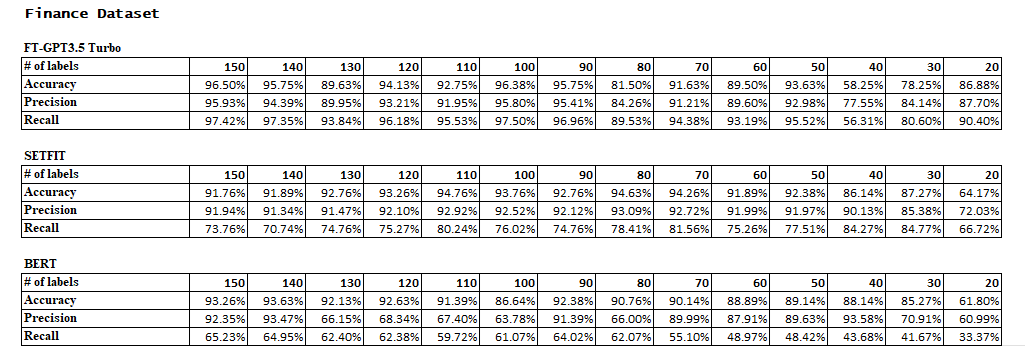}
\end{figure} 

\begin{figure}[h!]
\centering
\caption{Financial Phrasebank Dataset Plot}
\includegraphics[scale=0.4]{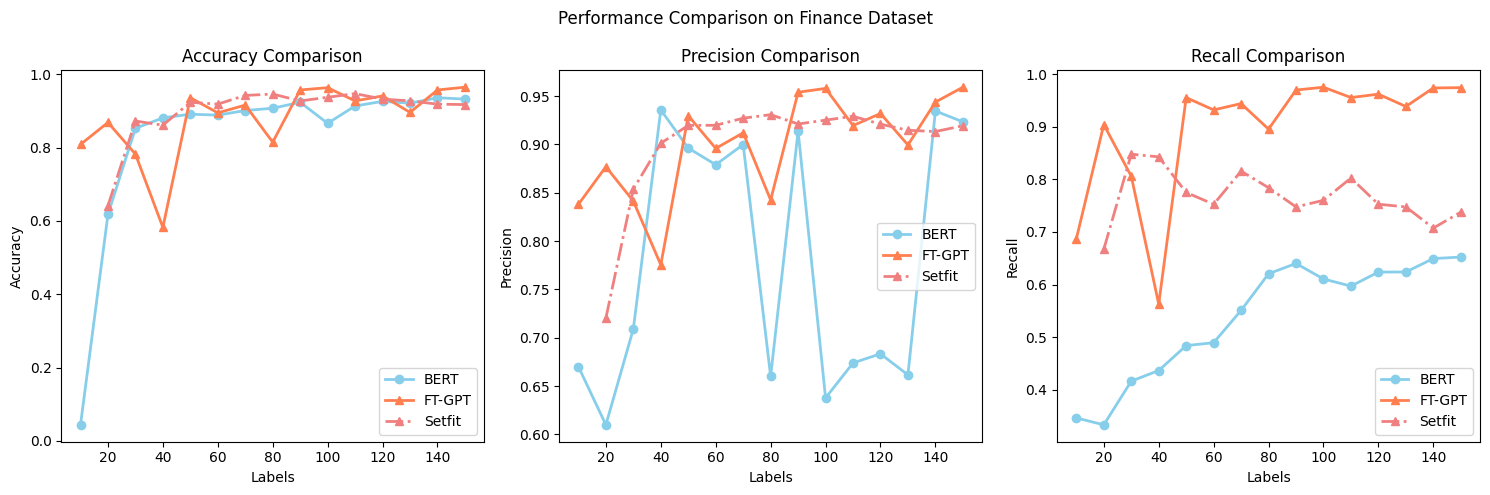}
\end{figure} 

\begin{figure}[h!]
\centering
\caption{Trec Coarse Label Dataset}
\includegraphics[scale=0.5]{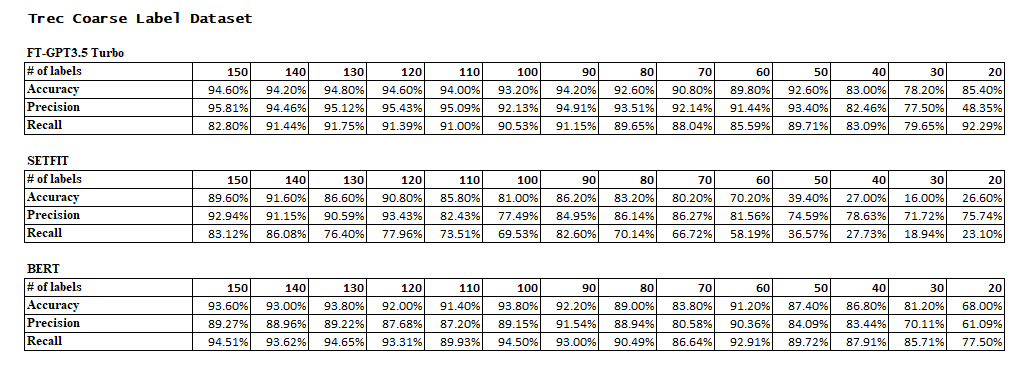}
\end{figure} 

\begin{figure}[h!]
\centering
\caption{Trec Coarse Dataset Plot}
\includegraphics[scale=0.4]{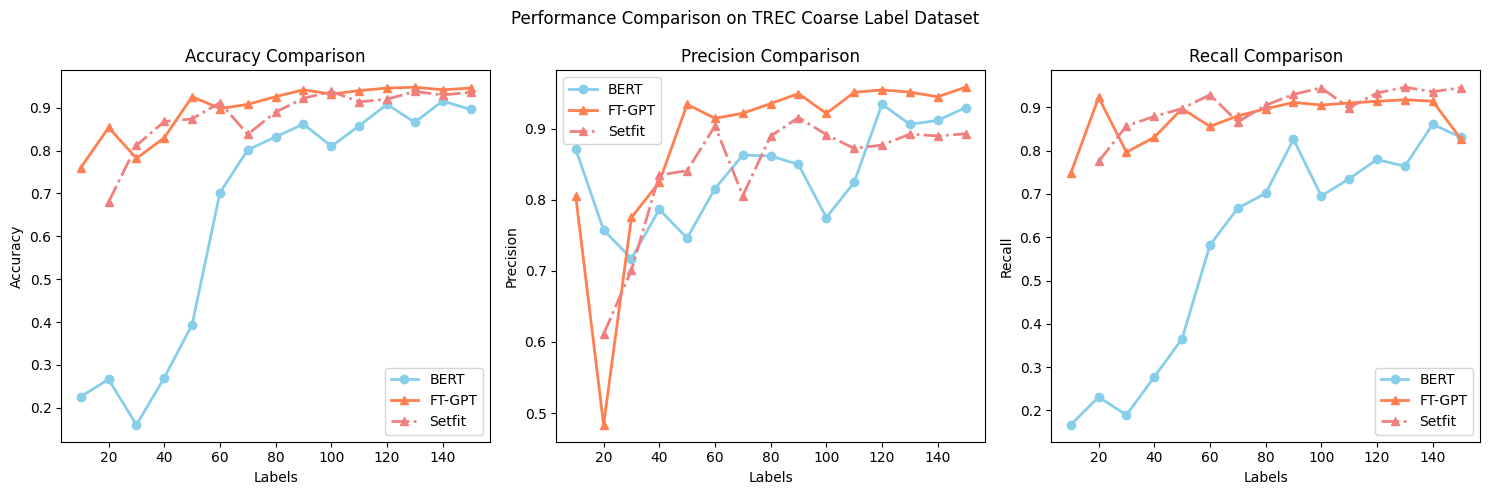}
\end{figure} 

\begin{figure}[h!]
\centering
\caption{Trec Fine Label Dataset}
\includegraphics[scale=0.5]{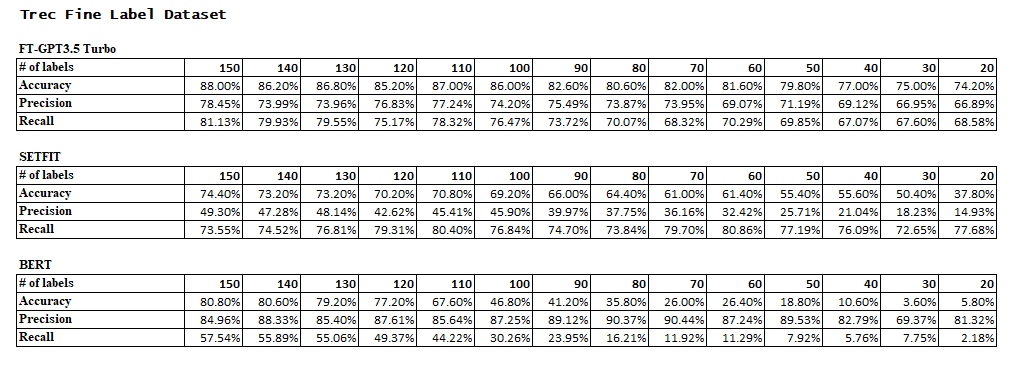}
\end{figure} 

\begin{figure}[h!]
\centering
\caption{Trec Fine Label Dataset Plot}
\includegraphics[scale=0.4]{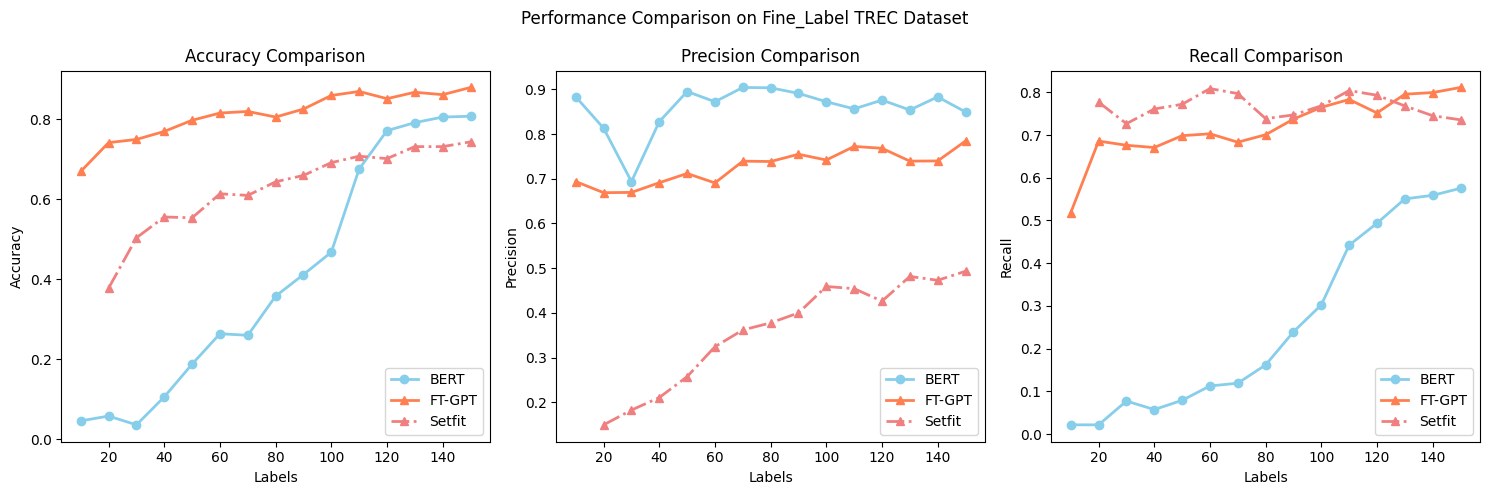}
\end{figure}

\clearpage
\section{Conclusion}
Our investigation into few-shot learning and active learning methodologies is promising in enhancing language models with minimal labeled data. We were able to produce notable improvements in model accuracy, recall, and precision across diverse domain-specific datasets by utilizing a continuous feedback loop and integrating human expertise. The ability to leverage a minimal number of labels to refine a model will be extremely beneficial to businesses. Companies will be able to maintain model performance levels while minimizing the resources typically required for manual labeling processes. While our research demonstrated the potential of few-shot learning with human feedback, there are some limitations. Our study is focused on a limited number of models such as GPT, BERT, and SetFit. To gain a more comprehensive understanding of the applicability of few-shot learning across various model structures, we can expand the work to train a diverse array of models. We could test newer model architectures like T5, Transformer-XL, or different BERT and GPT 4 variations.  Experimenting with additional models can provide deeper insights into the effectiveness of these approaches. 

\section{Next Steps} 
To address these limitations, follow-up research could include training the models on datasets with complex taxonomies and intricate label hierarchies. By employing datasets from domains such as medical, legal, or financial fields with complex taxonomies, we will challenge the model's ability to generalize. By looking at classification problems on millions of rows of data with hundreds to thousands of categories and subcategories, the accuracy of these models will be of significant importance for ROI in specific domains. This will establish the foundation for using complex, industry-specific data to train more accurate classification models, bridging the gap between model performance and real-world application.

\section{References}

% \printbibliography[heading=none]

\small
\begin{enumerate}
    
    \item What are Large Language Models (LLM)? \url{https://aws.amazon.com/what-is/large-language-model/}, Accessed on 2023-12-17.

    \item Chen Yanhui. A Battle Against Amnesia: A Brief History and Introduction of Recurrent Neural Networks. \url{https://towardsdatascience.com/a-battle-against-amnesia-a-brief-history-and-introduction-of-recurrent-neural-networks-50496aae6740}, Accessed on 2023-12-17.

    \item Kiel Dang. Language Model History — Before and After Transformer: The AI Revolution. \url{https://medium.com/@kirudang/language-model-history-before-and-after-transformer-the-ai-revolution-bedc7948a130}, Accessed on 2023-12-17.

    \item Sanchit Goel. Evolution of Transformers — Part 1. \url{https://sanchman21.medium.com/evolution-of-transformers-part-1-faac3f19d780}, Accessed on 2023-12-17.

    \item Liz McQuillan. Deep Learning 101: What Is a Transformer and Why Should I Care? \url{https://www.saltdatalabs.com/blog/deep-learning-101/what-is-a-transformer-and-why-should-i-care}, Accessed on 2023-12-17.

    \item Ashish Vaswani and Noam Shazeer and Niki Parmar and Jakob Uszkoreit and Llion Jones and Aidan N. Gomez and Lukasz Kaiser and Illia Polosukhin. Attention Is All You Need, 2023. arXiv:1706.03762.

    \item Shuohang Wang and Yang Liu and Yichong Xu and Chenguang Zhu and Michael Zeng. Want To Reduce Labeling Cost? GPT-3 Can Help, 2021. arXiv:2108.13487.

    \item Lewis Tunstall and Nils Reimers and Unso Eun Seo Jo and Luke Bates and Daniel Korat and Moshe Wasserblat and Oren Pereg. Efficient Few-Shot Learning Without Prompts, 2022. arXiv:2209.11055.

    \item Curtis Northcutt and Anish Athalye and Jonas Mueller. Label Errors in ML Test Sets. \url{https://labelerrors.com}, Accessed on 2023-12-23.

    \item Programmatic labeling. \url{https://snorkel.ai/programmatic-labeling/}, Accessed on 2023-12-23.

    \item Tom B. Brown and Benjamin Mann and Nick Ryder and Melanie Subbiah, et al. Language Models are Few-Shot Learners, 2020. arXiv:2005.14165.

    \item Romal Thoppilan, Daniel De Freitas, Jamie Hall, Noam Shazeer, Apoorv Kulshreshtha, et al. LaMDA: Language Models for Dialog Applications, 2022. arXiv:2201.08239.

    \item Aakanksha Chowdhery, Sharan Narang, Jacob Devlin, Maarten Bosma, Gaurav Mishra et al. PaLM: Scaling Language Modeling with Pathways, 2022. arXiv:2204.02311.

    \item Nan Du, Yanping Huang, Andrew M. Dai, Simon Tong, et al. GLaM: Efficient Scaling of Language Models with Mixture-of-Experts, 2022. arXiv:2112.06905.

    \item Jean-Baptiste Alayrac, Jeff Donahue, Pauline Luc, Antoine Miech, et al. Flamingo: a Visual Language Model for Few-Shot Learning, 2022. arXiv:2204.14198.

    \item Jack FitzGerald, Shankar Ananthakrishnan, Konstantine Arkoudas, Davide Bernardi, et al. Alexa Teacher Model: Pretraining and Distilling Multi-Billion-Parameter Encoders for Natural Language Understanding Systems. In Proceedings of the 28th ACM SIGKDD Conference on Knowledge Discovery and Data Mining (KDD ’22), August 2022. \url{http://dx.doi.org/10.1145/3534678.3539173}, DOI: 10.1145/3534678.3539173.

\end{enumerate}

\section{Appendix A: Dataset Descriptions}

\begin{enumerate}
\item Amazon Dataset
\begin{itemize}
    \item Description: Amazon product reviews
    \item Source: Aymeric Roucher on HuggingFace
    \item Size: 6000 rows
    \item URL: \url{https://huggingface.co/datasets/A-Roucher/amazon_product_reviews_datafiniti}
\end{itemize}
\item Finance Phrasebank Dataset
\begin{itemize}
    \item Description: Finance phrases from news along with a rating out of 5 of whether they are negative, neutral, or positive
    \item Source: HuggingFace
    \item Size: 4850 rows
    \item URL: \url{https://huggingface.co/datasets/financial_phrasebank}
\end{itemize}
\item TREC Dataset
\begin{itemize}
    \item Description: Text Retrieval Conference (TREC) with coarse labels and fine labels
    \item Source: HuggingFace
    \item Size: 5450 rows
    \item URL: \url{https://huggingface.co/datasets/trec}
\end{itemize}
\end{enumerate}

\end{document}